\documentclass[11pt,a4paper]{article}
\usepackage[whole]{bxcjkjatype} 
\usepackage[unicode]{hyperref}
\usepackage[hyperref]{acl2018}
\usepackage{times}
\usepackage{latexsym}

\usepackage{url}

\usepackage{graphicx}
\usepackage{amsmath}
\usepackage{amssymb}
\usepackage{multirow}
\usepackage{bm}
\usepackage{color}

\aclfinalcopy 
\def\aclpaperid{20} 

\title{Recursive Neural Network Based Preordering for English-to-Japanese Machine Translation}

\author{Yuki Kawara$^\dag$ \\\And
Chenhui Chu$^\ddag$ \\
$^\dag$Graduate School of Information Science and Technology, Osaka University \\
$^\ddag$Institute for Datability Science, Osaka University \\
{\tt \{kawara.yuki,arase\}@ist.osaka-u.ac.jp,chu@ids.osaka-u.ac.jp} \\\And
Yuki Arase$^\dag$ \\
}

\date{2018/5/21}

\begin{document}
\maketitle
\begin{abstract}
The word order between source and target languages significantly influences the translation quality in machine translation. Preordering can effectively address this problem.　Previous preordering methods require a manual feature design, making language dependent design costly.
In this paper, we propose a preordering method with a recursive neural network that learns features from raw inputs. Experiments show that the proposed method achieves comparable gain in translation quality to the state-of-the-art method but without a manual feature design.
\end{abstract}

\section{Introduction}
The word order between source and target languages significantly influences the translation quality in statistical machine translation (SMT) \cite{Tillmann:2004:UOM:1613984.1614010,hayashi-EtAl:2013:EMNLP,nakagawa2015}.
Models that adjust orders of translated phrases in decoding have been proposed to solve this problem \cite{Tillmann:2004:UOM:1613984.1614010, Koehnetal:2005edinburgh, Nagata:2006:CGP:1220175.1220265}.
However, such reordering models do not perform well for long-distance reordering. In addition, their computational costs are expensive.
To address these problems, preordering \cite{xia-mccord:2004:COLING, wang-collins-koehn:2007:EMNLP-CoNLL2007, xu-EtAl:2009:NAACLHLT09, isozaki10hfe, gojunfraser2012, nakagawa2015} and post-ordering \cite{goto-utiyama-sumita:2012:ACL2012short,Goto:2013:PPI:2523057.2518100,hayashi-EtAl:2013:EMNLP} models have been proposed. Preordering reorders source sentences before translation, while post-ordering reorders sentences translated without considering the word order after translation. In particular, preordering effectively improves the translation quality because it solves long-distance reordering and computational complexity issues \cite{jehl-EtAl:2014:EACL,nakagawa2015}.

Rule-based preordering methods either manually create reordering rules \cite{wang-collins-koehn:2007:EMNLP-CoNLL2007, xu-EtAl:2009:NAACLHLT09, isozaki10hfe, gojunfraser2012} or extract reordering rules from a corpus \cite{xia-mccord:2004:COLING, genzel:2010:PAPERS}. On the other hand, studies in \cite{neubig-watanabe-mori:2012:EMNLP-CoNLL, lerner-petrov:2013:EMNLP, hoshino-EtAl:2015:ACL-IJCNLP, nakagawa2015} apply machine learning to the preordering problem. \citet{hoshino-EtAl:2015:ACL-IJCNLP} proposed a method that learns whether child nodes should be swapped at each node of a syntax tree. \citet{neubig-watanabe-mori:2012:EMNLP-CoNLL} and \citet{nakagawa2015} proposed methods that construct a binary tree and reordering simultaneously from a source sentence. These methods require a manual feature design for every language pair, which makes language dependent design costly.
To overcome this challenge, methods based on feed forward neural networks that do not require a manual feature design have been proposed \cite{degispert-iglesias-byrne:2015:NAACL-HLT, botha-EtAl:2017:EMNLP2017}. However, these methods decide whether to reorder child nodes without considering the sub-trees, which contains important information for reordering.

As a preordering method that is free of manual feature design and makes use of information in sub-trees, we propose a preordering method with a recursive neural network (RvNN). RvNN calculates reordering in a bottom-up manner (from the leaf nodes to the root) on a source syntax tree. Thus, preordering is performed considering the entire sub-trees.
Specifically, RvNN learns whether to reorder nodes of a syntax tree\footnote{In this paper, we used binary syntax trees.}  with a vector representation of sub-trees and syntactic categories. We evaluate the proposed method for English-to-Japanese translations using both phrase-based SMT (PBSMT) and neural MT (NMT). The results confirm that the proposed method achieves comparable translation quality to the state-of-the-art preordering method \cite{nakagawa2015} that requires a manual feature design.

\section{Preordering with a Recursive Neural Network}
We explain our design of the RvNN to conduct preordering after describing how to obtain gold-standard labels for preordering.
\subsection{Gold-Standard Labels for Preordering}
\label{sec:rank_eval}
We created training data for preordering by labeling whether each node of the source-side syntax tree has reordered child nodes against a target-side sentence. The label is determined based on Kendall's $\tau$ \cite{kendall1938measure} as in \cite{nakagawa2015}, which is calculated by Equation (\ref{eq:kendall_tau}).
\begin{eqnarray}
\label{eq:kendall_tau}
\tau&=&\frac{4\sum^{|{\bf y}|-1}_{i=1}\sum^{|{\bf y}|}_{j=i+1} \delta ({\bf y}_i<{\bf y}_j)}{|{\bf y}|(|{\bf y}|-1)} - 1, \ \ \ \\
\delta(x)&=&\begin{cases}
1 & (x\ {\rm is\ true}), \\
0 & ({\rm otherwise}),
\end{cases} \nonumber
\end{eqnarray}
where ${\bf y}$ is a vector of target word indexes that are aligned with source words. The value of Kendall's $\tau$ is in $[-1,1]$. When it is $1$, it means the sequence of ${\bf y}$ is in a complete ascending order, {\em i.e.}, target sentence has the same word order with the source in terms of word alignment. At each node, if Kendall's $\tau$ increases by reordering child nodes, an ``Inverted'' label is assigned; otherwise, a ``Straight'' label, which means the child nodes do not need to be reordered, is assigned. When a source word of a child node does not have an alignment, a ``Straight'' label is assigned.

\subsection{Preordering Model}
\begin{figure}[t!]
\centering
\includegraphics[width=0.35\textwidth, bb=0 0 263 194]{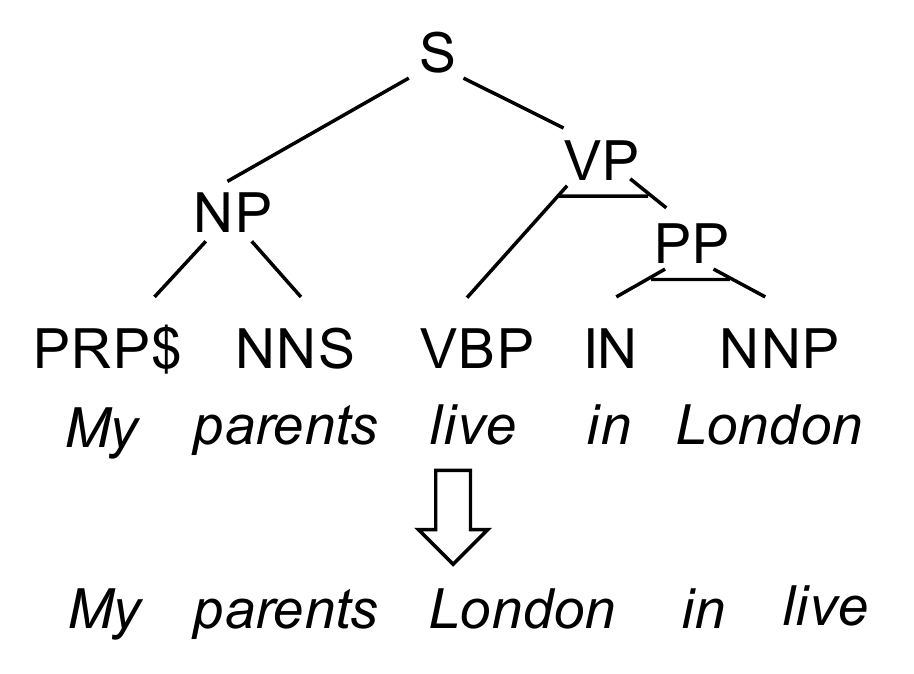}
\caption{Preordering an English sentence ``My parents live in London'' with RvNN (Nodes with a horizontal line mean ``Inverted'').}
\label{fig:rec_example}
\end{figure}

RvNN is constructed given a binary syntax tree. It predicts the label determined in Section \ref{sec:rank_eval} at each node. RvNN decides whether to reorder the child nodes by considering the sub-tree. The vector of the sub-tree is calculated in a bottom-up manner from the leaf nodes.
Figure \ref{fig:rec_example} shows an example of preordering of an English sentence ``My parents live in London.''
At the VP node corresponding to ``live in London,'' the vector of the node is calculated by Equation (\ref{eq:concatnodes}), considering its child nodes correspond to ``live'' and ``in London.''
\begin{eqnarray}
\label{eq:concatnodes}
{\bf p} &=& f([{\bf p_l}; {\bf p_r}]W + {\bf b}),\\
\label{eq:labelfunc}
{\bf s} &=& {\bf p}W_s + {\bf b}_s,
\end{eqnarray}
where $f$ is a rectifier, $W \in \mathbb{R}^{2\lambda \times \lambda}$ is a weight matrix, ${\bf p_l}$ and ${\bf p_r}$ are vector representations of the left and right child nodes, respectively. $[\cdot;\cdot]$ denotes the concatenation of two vectors. $W_s \in \mathbb{R}^{\lambda \times 2}$ is a weight matrix for the output layer, and ${\bf b}, {\bf b}_s \in \mathbb{R}^{\lambda}$ are the biases. ${\bf s}\in \mathbb{R}^2$ calculated by Equation (\ref{eq:labelfunc}) is a weight vector for each  label, which is fed into a softmax function to calculate the probabilities of the ``Straight'' and ``Inverted'' labels.

At a leaf node, a word embedding calculated by Equations (\ref{eq:embedding}) and (\ref{eq:leaf_node}) is fed into Equation (\ref{eq:concatnodes}).
\begin{eqnarray}
\label{eq:embedding}
{\bf e} &=& {\bm x}W_E, \\
\label{eq:leaf_node}
{\bf p_e} &=& f({\bf e}W_l + {\bf b}_l),
\end{eqnarray}
where ${\bm x} \in \mathbb{R}^{N}$ is a one-hot vector of an input word with a vocabulary size of $N$, $W_E \in \mathbb{R}^{N \times \lambda}$ is an embedding matrix, and ${\bf b}_l \in \mathbb{R}^\lambda$ is the bias.
The loss function is the cross entropy defined by Equation (\ref{eq:loss_func}).
\begin{eqnarray}
\label{eq:loss_func}
L(\theta) &=& -\frac{1}{K}\sum^K_{k=1} \sum_{n\in \mathcal{T}} l^n_k\log p(l^n_k; \theta),
\end{eqnarray}
where $\theta$ is the parameters of the model, $n$ is the node of a syntax tree $\mathcal{T}$, $K$ is a mini batch size, and $l^n_k$ is the label of the $n$-th node in the $k$-th syntax tree in the mini batch.

With the model using POS tags and syntactic categories, we use Equation (\ref{eq:concatnodes_pos}) instead of Equation (\ref{eq:concatnodes}).
 \begin{eqnarray}
\label{eq:concatnodes_pos}
{\bf p} &=& f([{\bf p_l}; {\bf p_r}; {\bf e}_t]W_t + {\bf b}_t),
\end{eqnarray}
where ${\bf e}_t$ represents a vector of POS tags or syntactic categories, $W_t \in \mathbb{R}^{3\lambda \times \lambda}$ is a weight matrix, and ${\bf b}_t \in \mathbb{R}^\lambda$ is the bias. ${\bf e}_t$ is calculated in the same manner as Equations (\ref{eq:embedding}) and (\ref{eq:leaf_node}), but the input is a one-hot vector of the POS tags or syntactic categories at each node. $\lambda$ is tuned on a development set, whose effects are investigated in Section \ref{sec:result}.

\section{Experiments}
\subsection{Settings}
We conducted English-to-Japanese translation experiments using the ASPEC corpus \cite{NAKAZAWA16.621}. This corpus provides $3$M sentence pairs as training data, $1,790$ sentence pairs as development data, and $1,812$ sentence pairs as test data.
We used Stanford CoreNLP\footnote{http://stanfordnlp.github.io/CoreNLP/} for tokenization and POS tagging, Enju\footnote{http://www.nactem.ac.uk/enju/} for parsing of English, and MeCab\footnote{http://taku910.github.io/mecab/} for tokenization of Japanese. For word alignment, we used MGIZA.\footnote{http://github.com/moses-smt/giza-pp} Source-to-target and target-to-source word alignments were calculated using IBM model $1$ and hidden Markov model, and they were combined with the intersection heuristic following \cite{nakagawa2015}.

We implemented our RvNN preordering model with Chainer.\footnote{http://chainer.org/} The ASPEC corpus was created using the sentence alignment method proposed in \citep{Utiyama07ajapanese-english} and was sorted based on the alignment confidence scores. In this paper, we used $100$k sentences sampled from the top $500$k sentences as training data for preordering. The vocabulary size $N$ was set to $50$k.
We used Adam \cite{DBLP:journals/corr/KingmaB14} with a weight decay and gradient clipping for optimization. The mini batch size $K$ was set to $500$.

We compared our model with the state-of-the-art preordering method proposed in \cite{nakagawa2015}, which is hereafter referred to as {\em BTG}. We used its publicly available implementation,\footnote{http://github.com/google/topdown-btg-preordering} and trained it on the same $100$k sentences as our model.

We used the $1.8$M source and target sentences as training data for MT. We excluded part of the sentence pairs whose lengths were longer than $50$ words or the source to target length ratio exceeded $9$.
For SMT, we used Moses.\footnote{http://www.statmt.org/moses/} We trained the $5$-gram language model on the target side of the training corpus with KenLM.\footnote{http://github.com/kpu/kenlm}
Tuning was performed by minimum error rate training \cite{och:2003:ACL}. We repeated tuning and testing of each model $3$ times and reported the average of scores.
For NMT, we used the attention-based encoder-decoder model of \cite{luong-pham-manning:2015:EMNLP} with $2$-layer LSTM implemented in OpenNMT.\footnote{http://opennmt.net/} The sizes of the vocabulary, word embedding, and hidden layer were set to $50$k, $500$, and $500$, respectively. The batch size was set to $64$, and the number of epochs was set to $13$. The translation quality was evaluated using BLEU \cite{papineni-EtAl:2002:ACL} and RIBES \cite{isozaki-EtAl:2010:EMNLP} using the bootstrap resampling method \cite{koehn:2004:EMNLP} for the significance test.

\begin{figure}[!t]
\centering
\includegraphics[width=0.48\textwidth, bb=0 0 720 576]{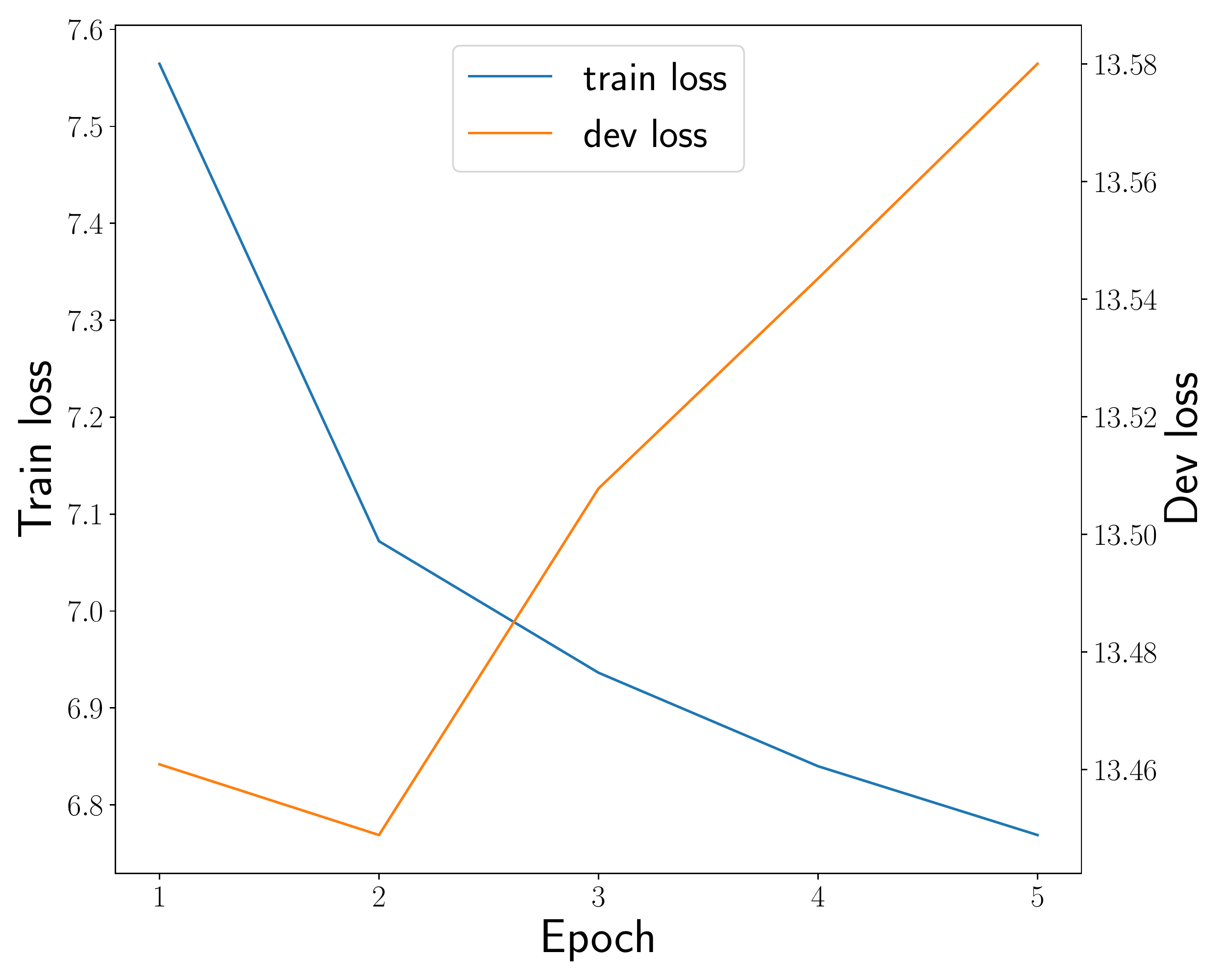}
\vspace{-8truemm}
\caption{Learning curve of our preordering model.}
\label{fig:loss_curve_enja}
\end{figure}

\begin{table}[t!]
\centering
\begin{tabular}{@{}c|c|c|c@{}}\hline
Node dimensions & $100$ & $200$ & $500$ \\\hline\hline
w/o preordering & \multicolumn{3}{c}{$22.73$} \\\hline
w/o tags and categories & $24.63$ & $24.95$ & $25.02$ \\\hline
w/ tags and categories& $25.22$ & $25.41$ & $25.38$ \\\hline
\end{tabular}
\caption{BLEU scores with preordering by our model and without preordering under different $\lambda$ settings (trained on a $500$k subset of the training data).}
\label{tab:rec_bleu}
\vspace{-3truemm}
\end{table}

\subsection{Results}
\setcounter{figure}{3}
\begin{figure*}[!t]
\centering
\includegraphics[width=0.75\textwidth, bb=0 0 802 217]{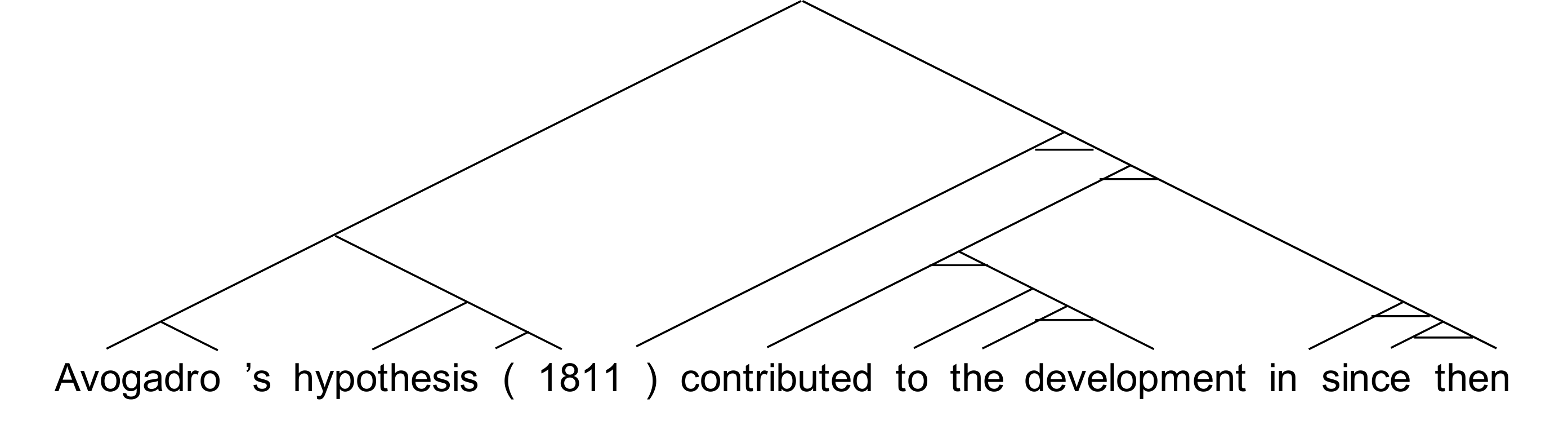}
\caption{Example of a syntax tree with a parse-error (the phrase ``(1811)'' was divided in two phrases by mistake). Our preordering result was affected by such parse-errors. (Nodes with a horizontal line means ``Inverted''.)}
\label{fig:parse_ex}
\end{figure*}
\label{sec:result}
Figure \ref{fig:loss_curve_enja} shows the learning curve of our preordering model with $\lambda=200$.\footnote{The learning curve behaves similarly for different $\lambda$ values.}
Both the training and the development losses decreased until $2$ epochs. However, the development loss started to increase after $3$ epochs. Therefore, the number of epochs was set up to $5$, and we chose the model with the lowest development loss. The source sentences in the translation evaluation were preordered with this model.

\begin{table}[t!]
\centering
{\small
\begin{tabular}{@{}l||c|c|c|c@{}}\hline
\multicolumn{1}{c||}{} & \multicolumn{2}{c|}{PBSMT} & \multicolumn{2}{c}{NMT} \\\hline
 & BLEU & RIBES & BLEU & RIBES \\\hline\hline
w/o preordering &  22.88 & 64.07 &  {\bf 32.68} &  {\bf 81.68} \\\hline
w/ BTG & {\bf 29.51} & {\bf 77.20} & 28.91 & {\bf 79.58} \\\hline
w/ RvNN & {\bf 29.16} & {\bf 76.39} & 29.01 & {\bf 79.63} \\\hline
\end{tabular}}
\caption{BLEU and RIBES scores on the test set. (All models are trained on the entire training corpus of $1.8$M sentence pairs.) Numbers in {\bf bold} indicate the best systems and the systems that are statistically insignificant at $p < 0.05$ from the best systems.}
\label{tab:eval_bleu}
\end{table}

\setcounter{figure}{2}
\begin{figure}[t!]
\centering
\includegraphics[width=0.48\textwidth, bb=0 0 460 345]{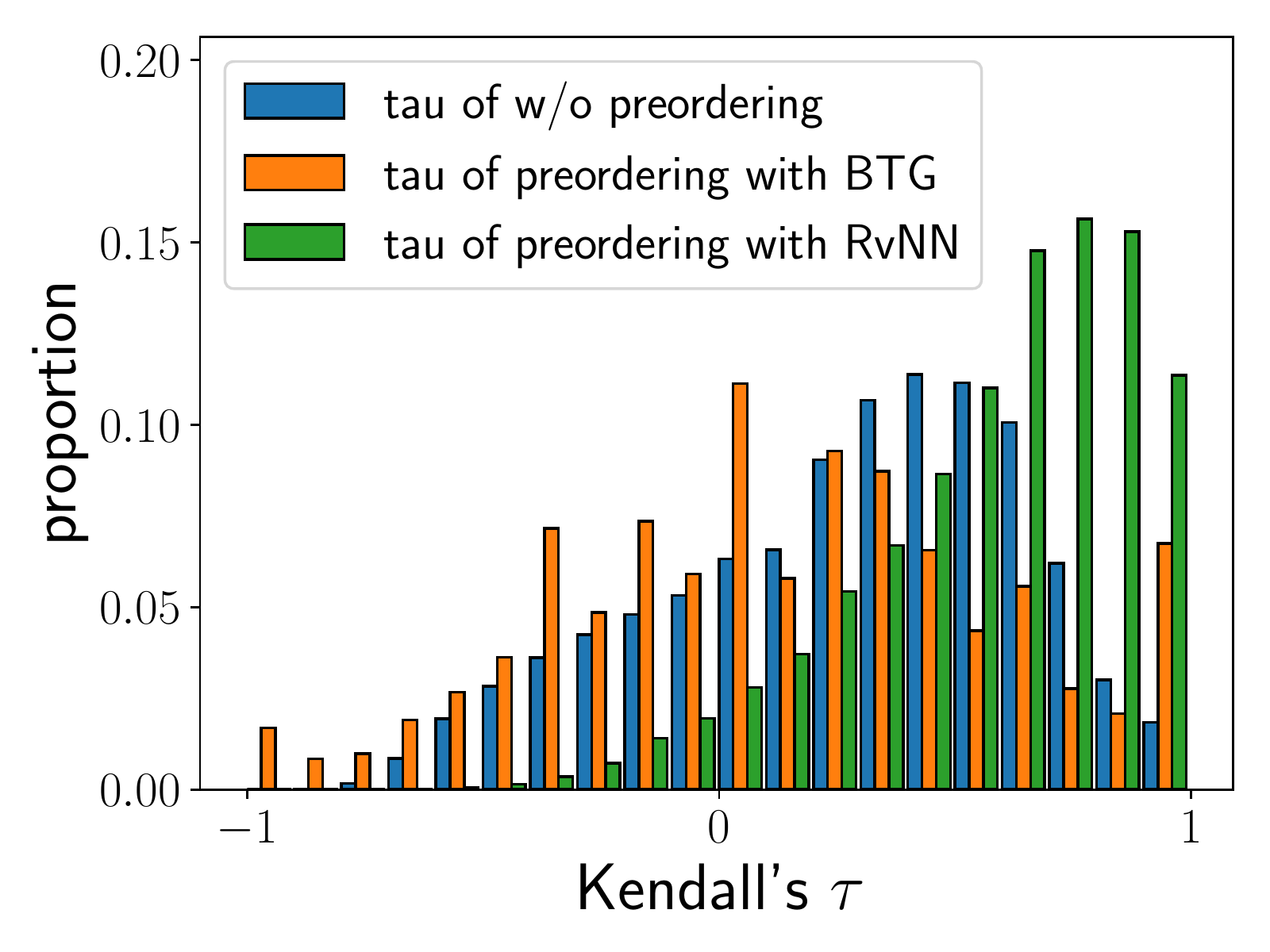}
\caption{Distribution of Kendall's $\tau$ in the training data without preordering, preordering by BTG, and preordering by our RvNN.}
\label{fig:tau_bleu}
\end{figure}
\setcounter{figure}{4}

\begin{table*}[t!]
\centering
{\small
\begin{tabular}{@{}l|l@{}}
\hline
\multicolumn{2}{c}{Preordered examples} \\\hline
Source sentence & because of the embedding heterostructure, current leakage around the threshold was minimal. \\
BTG & of the embedding heterostructure because, the threshold around current leakage minimal was. \\
RvNN & embedding heterostructure the of because, around threshold the current leakage minimal was. \\\hline\hline
\multicolumn{2}{c}{Translation examples by PBSMT} \\\hline
Reference & 埋込みヘテロ構造のため、しきい値近くでの漏れ電流は非常に小さかった。 \\
 & (embedding heterostructure of because, threshold around leakage very minimal.) \\
w/o preordering & 埋込みヘテロ構造のため、漏れ電流のしきい値付近では最低であった。 \\
 & (embedding heterostructure of because, leakage threshold around minimal.) \\
BTG　 & の埋込みヘテロ構造のため、このしきい値付近での漏れ電流の最小であった。 \\
 & (of embedding heterostructure of because, the threshold around leakage minimal.) \\
RvNN & 埋込みヘテロ構造のため、周辺のしきい値の電流漏れは認められなかった。 \\
 & (embedding heterostructure of because, around threshold leakage recognized not.) \\\hline
\end{tabular}}
\caption{Example where preordering improves translation. (Literal translations are given in the parenthesis under the Japanese sentences.)}
\label{tab:preorder_success}
\end{table*}

\begin{table*}[t!]
\centering
{\small
\begin{tabular}{l|l}
\hline
\multicolumn{2}{c}{Preordered examples} \\\hline
Source sentence & avogadro's hypothesis (1811) contributed to the development in since then. \\
BTG & avogadro's hypothesis (1811) the then since in development to contributed . \\
RvNN & avogadro's hypothesis (1811 then since in to development the contributed).  \\\hline\hline
\multicolumn{2}{c}{Translation examples by PBSMT} \\\hline
Reference & Avogadroの仮説 (1811)は，以後の発展に貢献した。 \\
& (Avogadro's hypothesis (1811), since then development to contributed.) \\
w/o preordering & Avogadroの仮説 (1811)の開発に貢献し以後である。 \\
& (Avogadro's hypothesis (1811) development to contributed since then.) \\
BTG & Avogadroの仮説 (1811)以後の発展に貢献した。 \\
& (Avogadro's hypothesis (1811) since then development to contributed.) \\
RvNN & Avogadroの仮説 (1811以降のこれらの開発に貢献した。　\\
& (Avogadro's hypothesis (1811 since then these development to contributed.) \\\hline
\end{tabular}}
\caption{Example of a parse-error disturbed preordering in our method. (Literal translations are given in the parenthesis under the Japanese sentences.)}
\label{tab:translation_ex}
\end{table*}

Next, we investigated the effect of $\lambda$. Table \ref{tab:rec_bleu} shows the BLEU scores with different $\lambda$ values, as well as the BLEU score without preordering. In this experiment, PBSMT was trained with a $500$k subset of training data, and the distortion limit was set to $6$. 
Our RvNNs consistently outperformed the plain PBSMT without preordering. The BLEU score improved as $\lambda$ increased when only word embedding was considered. In addition, RvNNs involving POS tags and syntactic categories achieved even higher BLEU scores. This result shows the effectiveness of POS tags and syntactic categories in reordering. For these models, setting $\lambda$ larger than $200$ did not contribute to the translation quality. Based on these, we further evaluated the RvNN with POS tags and syntactic categories where $\lambda=200$.

Table \ref{tab:eval_bleu} shows BLEU and RIBES scores of the test set on PBSMT and NMT trained on the entire training data of $1.8$M sentence pairs. The distortion limit of SMT systems trained using preordered sentences by RvNN and BTG was set to $0$, while that without preordering was set to $6$.
Compared to the plain PBSMT without preordering, both BLEU and RIBES increased significantly with preordering by RvNN and BTG. These scores were comparable (statistically insignificant at $p<0.05$)  between RvNN and BTG,\footnote{The $p$-value for BLEU and RIBES were $0.068$ and $0.226$, respectively.} indicating that the proposed method achieves a translation quality comparable to BTG.
In contrast to the case of PBSMT, NMT without preordering achieved a significantly higher BLEU score than NMT models with preordering by RvNN and BTG. This is the same phenomenon in the Chinese-to-Japanese translation experiment reported in \cite{sudoh-nagata:2016:WAT2016}. We assume that one reason is the isolation between preordering and NMT models, where both models are trained using independent optimization functions. 
In the future, we will investigate this problem and consider a model that unifies preordering and translation in a single model.

Figure \ref{fig:tau_bleu} shows the distribution of Kendall's $\tau$ in the original training data as well as the distributions after preordering by RvNN and BTG. The ratio of high Kendall's $\tau$ largely increased in the case of RvNN, suggesting that the proposed method learns preordering properly. Furthermore, the ratio of high Kendall's $\tau$ by RvNN is more than that of BTG, implying that preordering by RvNN is better than that by BTG.

We also manually investigated the preordering and translation results. We found that our model improved both.
Table \ref{tab:preorder_success} shows a successful preordering and translation example on PBSMT. 
The word order is notably different between source and reference sentences. 
After preordering, the word order between the source and reference sentences became the same.
Because RvNN depends on parsing, sentences with a parse-error tended to fail in preordering. For example, the phrase ``(1811)'' in Figure \ref{fig:parse_ex} was divided in two phrases by mistake. Consequently, preordering failed.
Table \ref{tab:translation_ex} shows preordering and translation examples for the sentence in Figure \ref{fig:parse_ex}.
Compared to the translation without preordering, the translation quality after preordering was improved to deliver correct meaning.

\section{Conclusion}
In this paper, we proposed a preordering method without a manual feature design for MT. The experiments confirmed that the proposed method achieved a translation quality comparable to the state-of-the-art preordering method that requires a manual feature design. As a future work, we plan to develop a model that jointly parses and preorders a source sentence. In addition, we plan to integrate preordering into the NMT model.

\section*{Acknowledgement}
\noindent This work was supported by NTT communication science laboratories and Grant-in-Aid for Research Activity Start-up \#17H06822, JSPS. 

\bibliography{acl2018}
\bibliographystyle{acl_natbib}

\end{document}